# AnoDFDNet: A Deep Feature Difference Network for Anomaly Detection

Zhixue Wang, *Yu Zhang*\*, Lin Luo, Nan Wang

*Abstract*—This paper proposed a novel anomaly detection (AD) approach of High-speed Train images based on convolutional neural networks and the Vision Transformer. Different from previous AD works, in which anomalies are identified with a single image using classification, segmentation, or object detection methods, the proposed method detects abnormal difference between two images taken at different times of the same region. In other words, we cast anomaly detection problem with a single image into a difference detection problem with two images. The core idea of the proposed method is that the 'anomaly' usually represents an abnormal state instead of a specific object, and this state should be identified by a pair of images. In addition, we introduced a deep feature difference AD network (AnoDFDNet) which sufficiently explored the potential of the Vision Transformer and convolutional neural networks. To verify the effectiveness of the proposed AnoDFDNet, we collected three datasets, a difference dataset (Diff Dataset), a foreign body dataset (FB Dataset), and an oil leakage dataset (OL Dataset). Experimental results on above datasets demonstrate the superiority of proposed method. Source code are available at https://github.com/wangle53/AnoDFDNet.

*Index Terms*—Anomaly Detection, Difference Detection, Vision Transformer, Convolutional Neural Networks.

## I. Introduction

ANOMALY detection (AD) is one of the core tasks in computer vision which has been well-studied within a wide range of research areas and application domains. Its critical idea is how to identify abnormal information that significantly deviate from the majority of data information [1, 2]. Depending on different application situation or data types, an anomaly also knows as outlier or novelty [3]. Approaches and applications of anomaly detection exist in various domains, such as fraud detection [4], video surveillance [5, 6], health-care [7], security check [8], fault detection [17, 18], defect detection [9-16, 19-21], and so on.

In the field of safety inspection of high-speed trains, AD is widely used to identify defects and anomalies. Due to the extreme importance of train safety inspection in ensuring the safe and reliable operation of high-speed trains, and the development of machine vision, more and more applications based on AD task are used in train safety inspection to improve detection efficiency, reduce detecting coast and realize intelligent detection [11-16]. In these methods, most of them focus on how to detect anomalies and identify surface defects of key components. Currently, inspired by the success of Convolutional Neural Networks (CNNs) on image analysis, more and more AD algorithms are equipped with CNNs to meet the requirement of fast computing speed, high efficiency and detecting accuracy [19-28]. According to the image analysis types, the popular AD methods for high-speed train safety inspection can be divided into four categories, unsupervised generated methods [7-10], anomaly or defect classification [16, 23, 26, 32-34], abnormal object detection [17-22, 24, 27], and defect segmentation [29-31, 33].

Even though above methods achieved considerable performance in detecting anomalies, we think using object-based methods to detect anomalies is not a very appropriate way. The main reason lies in three respects:

(1) The 'anomaly' usually represents an abnormal state instead of a specific object. As shown in Fig. 1, we illustrated some abnormal samples taken from the bottom of high-speed trains. For sample (a) (d) (f), and sample (c), it is available that using an object detection or segmentation model to detect 'foreign body' and 'scratch'. In this case, there remain a specific abnormal object to detect. However, for sample (b) and (e), since the anomaly is caused by a missed component instead of an added abnormal object, object detection or segmentation method doesn't work on it. In other words, there don't exist a specific abnormal object which can be localized or identified. Therefore, the object-based detecting methods are useless to identify anomalies if there are no abnormal objects.

(2) Each object usually has obvious features that represent itself belonging to a category. In AD tasks, different abnormal categories have different abnormal objects and some of the objects in each abnormal category have very different feature representation. As sample (a) (d) (f) shown in Fig. 1, the category 'foreign body' has different object types, such as tissue, packaging tape, or other unlisted abnormal objects. These abnormal objects have different sizes, shapes, and less of structure features. It would make a pre-trained model be difficult to detect unseen abnormal objects that their geometric appearances are very different from training data. In addition, one thing we can't ignore is that, due to the lack of abnormal

This work was supported in part by the National Natural Science Foundation of China (grant numbers 61771409), the Science and Technology Program of Sichuan (grant numbers 2021YJ0080).

Zhixue. Wang, Lin. Luo and Nan. Wang are with the School of Physical Science and Technology, Southwest Jiaotong University (e-mail: wangzhixue68@163.com).

Yu. Zhang, is with the School of Physical Science and Technology, Southwest Jiaotong University (e-mail: zhang.yuer@163.com).

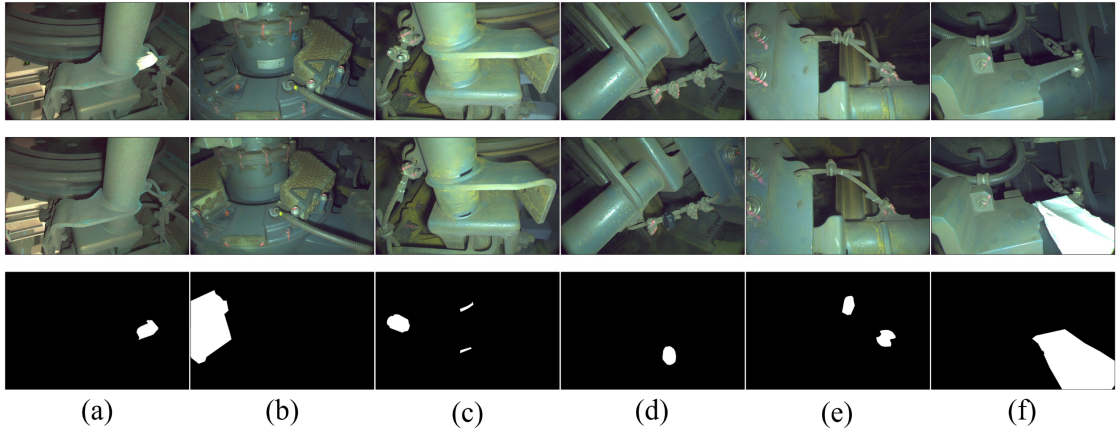

Fig. 1. Typical examples of Diff dataset (Difference Dataset). From up to down, the first, second, and third row denote the image taken at t0 (we denote as 'cur' image), the image taken at t1 (we denote as 'his' image), and the ground truth. We list some examples with typical anomalies. Sample (a) (d) and (f) denote 'foreign body'; Sample (b) and (e) denote 'component missing'; Sample (c) denotes 'scratch';

samples, it is very difficult to train a powerful enough model that is able to detect all of potential abnormal objects.

(3) During the train operation, there might have other foreign bodies, such as cables, tree branches, the bodies of animals or birds, stones, plastic bags, mud or ice and snow which may threaten train operation safety, and so on. It is impossible to list all of anomaly categories and abnormal object types. Therefore, there is no doubt that a pre-trained model would fail to detect unseen anomalies and abnormal objects. For example, a 'foreign body' dataset contains samples of stones, plastic bags, and tissues, the pre-trained model will have the capacity of detecting these abnormal objects. For the unseen foreign bodies such as tree branches, animal or bird bodies, as they didn't be trained by the model and have different representation features from trained objects, they would be miss-detected.

In this paper, we proposed a novel AD method. Different from previous methods, we use two images taken at different times to detect anomalies by differentiating their features. To the best of knowledge, the use of paired images and Siamese architectures for AD task is an area of research that has been studied only insufficiently. As shown in Fig. 2, a pair of samples contain two images taken at different times of the same region of the same train. For convenience, the image acquired at previous time is denoted as the 'history image', and the image acquired at the latest time is denoted as the 'current image'. As we mentioned above, the 'anomaly' describes a state instead of a specific object, and some anomalies can't be identified only from a single image. Therefore, it is much easier to identify whether there exists 'anomaly' using two images' comparison than only using a single image [26]. No matter what type the abnormal object is, there must remain difference between the normal and abnormal images. This difference denotes the abnormal information. At this point AD task is turned into a problem of detecting whether a pair of samples exist interesting differences.

The overview of the proposed model can be seen from Fig. 2. This model consists of four components: two CNNs used to extract the input image features; the Transformer-based module used to establish long-range relationships; the deep feature difference decoder used to generate anomaly maps.

The contribution can be summarized as follows:

(1) This paper proposed a novel AD method that accepts two images as input to detect anomalies, which casts abnormal object detection problem to a difference detection problem. Turn the problem of inexhaustibility of anomalies into a dichotomous problem of whether there exists difference.

(2) A deep feature difference AD model named AnoDFDNet is designed to deliver AD task, which sufficiently explored the potential of Vision Transformer and CNNs.

(3) We collected three AD datasets of high-speed train to evaluate the effectiveness of the proposed AnoDFDNet, a difference detection dataset (Diff Dataset), a foreign body dataset (FB Dataset), and an oil leakage dataset (OL dataset). The proposed AnoDFDNet achieved 76.24%, 81.04%, and 83.92% in terms of the F1 score, respectively.

The rest of this paper is organized as follows: Section II explores other works that were used for inspiration or comparison during the development of this work. The proposed method is described in Section III. The datasets and experiment configuration are described in Section IV. Section V shows the experimental results. Finally, the conclusion of this paper is drawn in Section VI.

## II. RELATED WORKS

Recently, inspired by the high performance that CNNs have achieved in image tasks, many CNNs-based methods have been introduced in AD task. In view of the lack of abnormal data, the unsupervised methods firstly train a model only on samples considered to be normal and then identify insufficiently abnormal samples which differ from the learned data distribution of normal data [7, 8]. Salehi et al. [9] used knowledge distillation to learn a cloner network from an expert network. This method detects and localizes anomalies using the discrepancy between the expert and cloner networks' intermediate activation values. Bergmann et al. [10] proposed a student-teacher anomaly detection model. It detects anomalies according to the output difference between the student network and the teacher network. For anomaly classification, the task comes to directly classifying different types of anomalies or defects. Wang et al. [26] proposed three kinds of classification models to classify paired images. In their dataset, different anomalies are divided into six categories. As the same as [26], Gibert et al. [33] represented a classification method to identify different material of railway track images. For abnormal object detection, it usually takes two steps: first localizing the key component or the region needing to be detected, second using a classifier to identify whether this region or key component is abnormal or which category it belongs to. Chen et al. [19] proposed a coarse-to-fine detecting method. Their networks consist of three sub-networks, two detectors sequentially localize the cantilever joints and their fasteners, a classifier to diagnose the fasteners' defects. According to different defect types, the fasteners are classified into three categories, 'normal', 'missing' and 'latent missing'. Tu et al. [24] proposed a real-time defect detection method of track components. In this method, they take the influence of sample imbalance and subtle difference among different classes into consideration. Segmentation-based methods of AD task are usually used to segment defects. Further process is using a classifier to classify the segmented defects into different types. Cao et al. [30] proposed a pixel-level segmentation network for surface defect detection. To improve the model performance, the authors integrated feature aggregation and attention modules into the network. Their experiments demonstrated this model achieved good performance. Li et al. [31] presented a semantic-segmentation-based algorithm for the state recognition of rail fasteners. In addition, the pyramid scene analysis network and vector geometry measurements are combined into the network to further improve its performance.

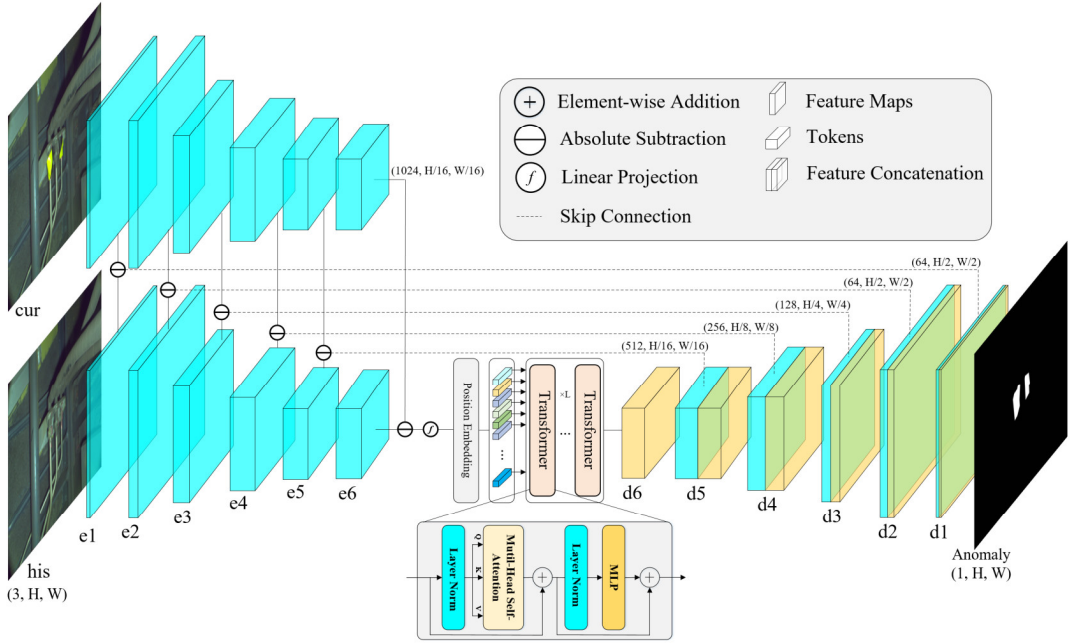

Fig. 2. Illustration of the proposed AnoDFDNet.

From the above works, it can be observed that the most existing approaches often focus on detecting specific anomalies. Therefore, these works cannot be applied to unseen anomalies which didn't be studied by the network. As mentioned in Section I, this is the motivation why we designed the proposed method. Since all anomalies can be identified by two image comparison, using a feature difference network can overcome above drawbacks and regardless of the specific categories of anomalies.

## III. METHOD

The overview architecture of the proposed AnoDFDNet is presented in Fig. 2. This model consists of four parts: two weights-shared CNNs used to extracted image features; a Vision Transformer-based [35-37] module used to establish long-range relationships and reform feature representation; the deep feature difference decoder used to fuse differentiated features and generate final detecting results. The corresponding layers between CNNs and the decoder are connected using skip connection.

Given two images $cur$, $his \in \mathbb{R}^{H \times W \times C}$ acquired at different times of the same region of the same train, where $H \times W$ denotes its spatial resolution, $C$ denotes the channel dimension. First, input images are fed into CNNs to extracted feature representation:

$$F_c = \text{CNN}(cur); F_h = \text{CNN}(his); \ F_c, F_h \in \mathbb{R}^{h \times w \times c}. \quad (1)$$

Before feeding feature representation into Transformer, we calculate their feature difference using absolute subtraction operation.

$$F = abs(F_c - F_h). \quad (2)$$

The differentiated features $F \in \mathbb{R}^{h \times w \times c}$ are converted into token sequences $p_i \in \mathbb{R}^c$ using transpose, reshape operation.

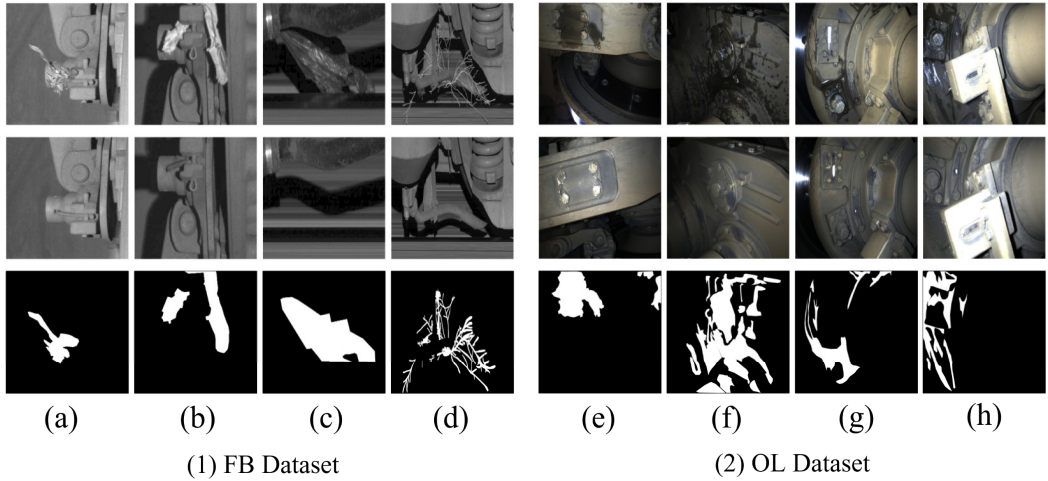

Fig. 3. Typical samples of FB dataset and OL Dataset. From up to down, each row denotes the 'cur' image, the 'his' image, and the ground truth, respectively.

Then, we map them into a latent d-dimensional embedding space using a trainable linear projection. After that, a learnable position embedding $E_{pos} \in \mathbb{R}^{l \times d}$ which is able to retain positional information is added to token sequences.

$$z_0 = [E(p_1); E(p_2); E(p_3); ...; E(p_l)] + E_{pos}, \quad (3)$$

where $l = h \times w$ presents the length of token sequences, $E(p_i) \in \mathbb{R}^d$ denotes the ith token sequence after linear projection.

The proposed AnoDFDNet consists of L layers of Transformer. As shown in Fig. 2, each Transformer is stocked with a multi-head self-attention (MSA) block, a multi-layer perceptron (MLP) block, and two layer-normalization (LN) layers. At each layer $\ell$, its output can be described as follows:

$$z'_\ell = \text{MSA}(\text{LN}(z_{\ell-1})) + z_{\ell-1}, \quad \ell = 1 \cdots L, \quad (4)$$

$$z_\ell = \text{MLP}(\text{LN}(z'_\ell)) + z'_\ell, \quad \ell = 1 \cdots L. \quad (5)$$

The decoder followed by Transformer layers is used to fuse the output features generated from the last layer and differentiated features from the previous layers.

**Loss Function:** Binary Cross Entropy (BCE) loss is used to optimized the proposed model. For the generated anomaly map $\hat{A}$, $a_{i,j} \in [0,1]$ denotes the prediction scores of a pixel on the anomaly map, $y_{i,j} \in \{0,1\}$ represents the pixel label at (i, j), where $y_{i,j} = 0$ denotes it is a normal sample, and $y_{i,j} = 1$ denotes it is an abnormal sample. The loss function can be formulated as follows:

$$Loss = \frac{1}{H \times W} \sum_i^H \sum_j^W -y_{i,j} \log a_{i,j} + (1 - y_{i,j}) \times \log(1 - a_{i,j}). \quad (6)$$

## IV. EXPERIMENT IMPLEMENTATION DETAILS

### A. Datasets

To evaluate the performance of the proposed AnoDFDNet, we collected three challenging anomaly detection datasets, a difference detection dataset (Diff Dataset), a foreign body dataset (FB Dataset), and an oil leakage dataset (OL Dataset).

**Diff Dataset:** As shown in Fig. 1, this dataset contains 399 pairs of training and validation samples and 101 pairs of testing samples. All images are captured automatically by a robot with a camera. By walking under the high-speed train, it can take images of specific regions and components. The main anomalies of this dataset are foreign bodies, missing components, loose components, detached components, scratches, and other anomalies. The spatial resolutions of each image are all 1920×1200. Training dataset is split into training and validation sets with the ratio of 8:2. All images are scaled to 256×256 before being fed into network.

**FB Dataset:** This dataset is a collection of foreign body anomalies, and contains 180 pairs of training and validation samples and 39 pairs of testing samples, some of the samples are shown in Fig. 3 (1). Their spatial resolutions vary from 123×271 to 4247×1282. To be the same with Diff dataset's operation, the dataset is split into training and validation with the ratio of 8:2, and all images are scaled to 256×256.

**OL Dataset:** This dataset is a collection of oil leakage anomalies, and some samples are shown in Fig. 3 (2). It consists of 1275 pairs of training and validation samples and 153 pairs of testing samples with the same resolution of 2048×1536. Follow the same operation of above two datasets, the dataset is split into training and validation with the ratio of 8:2, and all images are scaled to 256×256.

It worth noting that all of above datasets have many noisy differences, such as illumination variation, component rotation, viewpoint variation, and so on. These noisy differences will have a certain impact on the anomaly detection. For an AD model, the critical idea lies in if it can detect true anomaly difference while rejecting noisy differences.

### B. Optimization and Evaluation

In our experiments, all networks were trained using the Adam algorithm with a learning rate of 2e-4. All experiments were implemented using Pytorch 1.10.0 and with a Nvidia

TABLE I
COMPARISON WITH POPULAR METHODS.

| Datasets | No | Methods | Pre | Re | OA | F1 | IoU |
|---|---|---|---|---|---|---|---|
| Diff Dataset | 1 | Unet [38] | 0.4867 | 0.1021 | 0.9878 | 0.1688 | 0.0922 |
| | 2 | Unet-cat [38] | 0.3107 | 0.2707 | 0.9838 | 0.2893 | 0.1691 |
| | 3 | DeeplabV3+ [39] | 0.3834 | 0.2674 | 0.9863 | 0.3151 | 0.1870 |
| | 4 | DeeplabV3+-cat [39] | 0.3075 | 0.6135 | 0.9785 | 0.4097 | 0.2576 |
| | 5 | FCLNet [40] | 0.4938 | 0.5075 | 0.9881 | 0.5006 | 0.3338 |
| | 6 | AnoDFDNet | **0.7416** | **0.7584** | **0.9940** | **0.7499** | **0.5999** |
| FB Dataset | 7 | Unet [38] | 0.5667 | 0.5617 | 0.8572 | 0.5642 | 0.3930 |
| | 8 | Unet-cat [38] | 0.7975 | 0.6452 | 0.9146 | 0.7133 | 0.5544 |
| | 9 | DeeplabV3+ [39] | 0.7624 | 0.5173 | 0.8940 | 0.6164 | 0.4455 |
| | 10 | DeeplabV3+-cat [39] | 0.6855 | 0.8604 | 0.9120 | 0.7630 | 0.6169 |
| | 11 | FCLNet [40] | 0.7777 | 0.7151 | 0.9206 | 0.7451 | 0.5937 |
| | 12 | AnoDFDNet | **0.8328** | **0.7891** | **0.9392** | **0.8104** | **0.6812** |
| OL Dataset | 13 | Unet [38] | 0.8118 | 0.7589 | 0.9518 | 0.7845 | 0.6454 |
| | 14 | Unet-cat [38] | 0.8022 | 0.6111 | 0.9376 | 0.6937 | 0.5310 |
| | 15 | DeeplabV3+ [39] | 0.7674 | **0.8361** | 0.9507 | 0.8003 | 0.6671 |
| | 16 | DeeplabV3+-cat [39] | 0.8086 | 0.7921 | 0.9533 | 0.8003 | 0.6671 |
| | 17 | FCLNet [40] | 0.7951 | 0.8250 | 0.9563 | 0.8098 | 0.6804 |
| | 18 | AnoDFDNet | **0.8395** | 0.8227 | **0.9613** | **0.8310** | **0.7109** |

RTX2070 GPU of 8G memory and Intel i7-8700 CPU. To evaluate the performance of the proposed AnoDFDNet, five evaluation measures are used, the precision (Pre), recall (Re), overall accuracy (OA), F1 score (F1), and intersection over union (IoU).

## V. EXPERIMENTS

### A. Comparison Experiments

In this section, we compared the proposed model with several popular methods which can be used to detect anomalies of paired images. To the best of our knowledge, it doesn't receive much attention on detecting anomalies using paired images. Therefore, we compared AnoDFDNet with two popular segmentation methods, Unet [38] and DeeplabV3+ [39]. In our view, the purpose of AD task is to generate an anomaly map, it is similar to image segmentation task except the inputs are paired images. Note that the inputs are two images, the paired images are concatenated into a 6-channel image (In Table I, '_cat' denotes two images are concatenated into a 6-channel image as input) before being fed into segmentation networks. In addition, since change detection is a task that can also process bi-temporal images, we selected FCLNet [40] as a comparison. In the following, if no otherwise specified, the claimed AnoDFDNet denotes the one with 2 Transformer layers.

As shown in Table I, comparison experiments were demonstrated on three datasets. It can be observed that the proposed AnoDFDNet achieved superior performance over all datasets. On Diff dataset, the proposed model achieved 74.99% and 59.99% in terms of the F1-score and IoU, respectively. Compared with FCL, AnoDFDNet has improvements of 24.93% and 26.61% on the F1-score and IoU. On FB dataset, AnoDFDNet obtained 81.04% and 68.12% in terms of the F1-score and IoU, and improved the performance by 6.53% and 8.75%. For OL dataset, the proposed model still achieved the superior performance with 83.10% in F1-score and 71.09% in IoU, which improved by 2.12% and 3.05%. On both of Diff and FB datasets, the performances of Unet_cat and Deeplabv3+_cat are better than Unet and Deeplabv3+. There are three main reasons for this, one is two images provide more information for feature extraction; more importantly, the difference information only can be detected using two images acquired at different times; the third one is that anomaly objects are inexhaustible and can't be listed in their entirety; for some specific anomalies, such as foreign bodies, they don't have a particular geometrical characteristic. Therefore, it is difficult to detect anomalies using a single image.

Another important point here is that the overall performance of OL dataset is higher than Diff and FB datasets. The main reason is that Diff and FB dataset have larger viewpoint than OL dataset. As shown in Fig. 4, we fused 'cur' and 'his' image into an image. As the existence of viewpoint difference, two images are unable to match with each other. In the other word, the two images are un-registered. This un-registered viewpoint difference makes detecting anomalies be more difficult. From Table I, on Diff dataset, the proposed AnoDFDNet improved with a huge margin compared with FCLNet. It indicates than AnoDFDnet is good to overcome viewpoint difference.

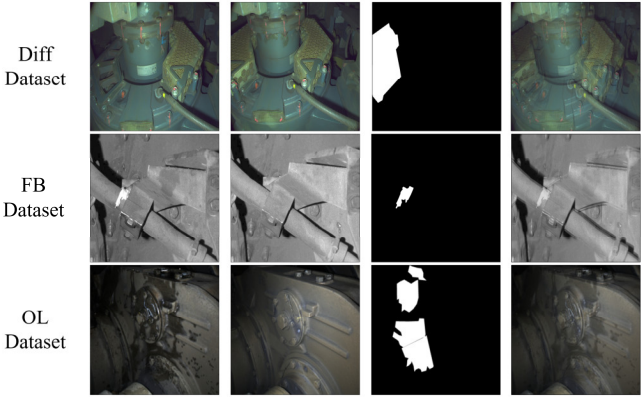

Fig. 4. Typical samples with large viewpoint difference. From left to right in each row, they are the 'cur' image, the 'his' image, the ground truth, and the fusion of both images, respectively.

TABLE II
COMPARISON OF INFERENCE SPEED (FRAME PER SECOND, FPS)

| No | Methods | Layer Number | FPS (CPU) | FPS (GPU) |
|---|---|---|---|---|
| 1 | Unet [38] | - | **8.61** | 102.14 |
| 2 | DeeplabV3+ [39] | - | 3.65 | 48.47 |
| 3 | FCLNet [40] | - | 1.15 | 50.49 |
| 4 | AnoDFDNet | 0 | 6.86 | **124.60** |
| 5 | AnoDFDNet | 1 | 5.51 | 95.81 |
| 6 | AnoDFDNet | 2 | 4.62 | 88.36 |
| 7 | AnoDFDNet | 4 | 3.48 | 78.35 |
| 8 | AnoDFDNet | 8 | 2.40 | 63.62 |

For further comparison, we figured out their receiver operating characteristic (ROC) curves. As shown in Fig. 5, the proposed model obtained the highest AUC (area under curve) and the lowest EER (equal error rate). AnoDFDNet achieved 98.61%, 96.80%, and 98.73% in terms of AUC, and 5.05%, 9.72%, and 5.46% in terms of EER, on Diff, FB, and OL dataset respectively. Table II shows the inference speeds of the methods, we can observe that the proposed method has a good trade-off between detecting accuracy and computing complexity. From above comparison, it indicates the proposed model has better capacity of detecting anomalies than other methods.

### B. Visualization

In this section, we have visualized anomaly maps of the selected methods to visually compare them. As shown in Fig. 6, on Diff dataset, it is no doubt that the presented AnoDFDNet performed the best. As the existence of large viewpoint difference between two images, all selected methods seem incompetent to detect anomalies. For example, for the fourth pair of samples, all selected methods missed the missing anomaly and the scratch anomaly; for the third pair of samples, Unet_cat detected the un-registered pipe fitting as anomalies instead of the foreign body. As to FB dataset, although its viewpoint difference is smaller than Diff dataset, it still brings difficulties to anomaly detection. For the second pair of samples, both Unet and Deeplabv3+ missed the foreign body anomaly, Unet_cat and Deeplabv3+_cat just detected a little part of the anomaly. With regard to OL dataset, even though the performances of all methods are comparable, some difference can be found on closer observation. Unet missed the anomalies on the top left corner of the first pair of samples. For the third pair of samples, the anomaly map detected by FCLNet is not fine enough. In the last pair of samples, only our model did a good job of detecting the oil leakage anomaly in the box.

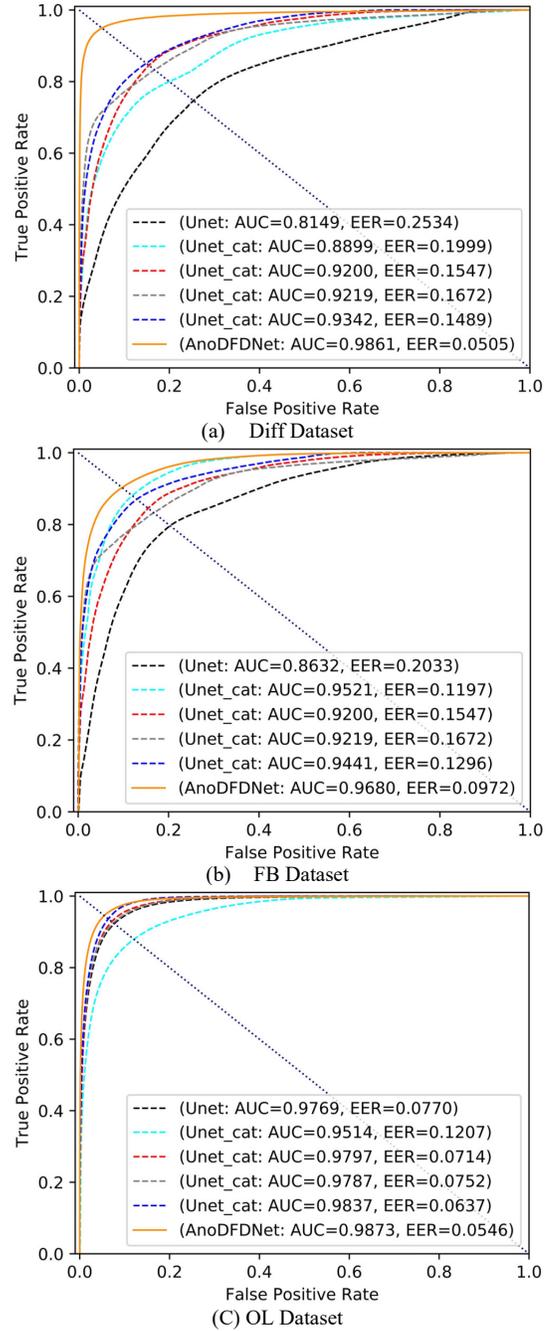

Fig. 5. Receiver operating characteristic (ROC) curves of different methods on three datasets, where AUC denotes the area under the curve, EER denotes the equal error rate.

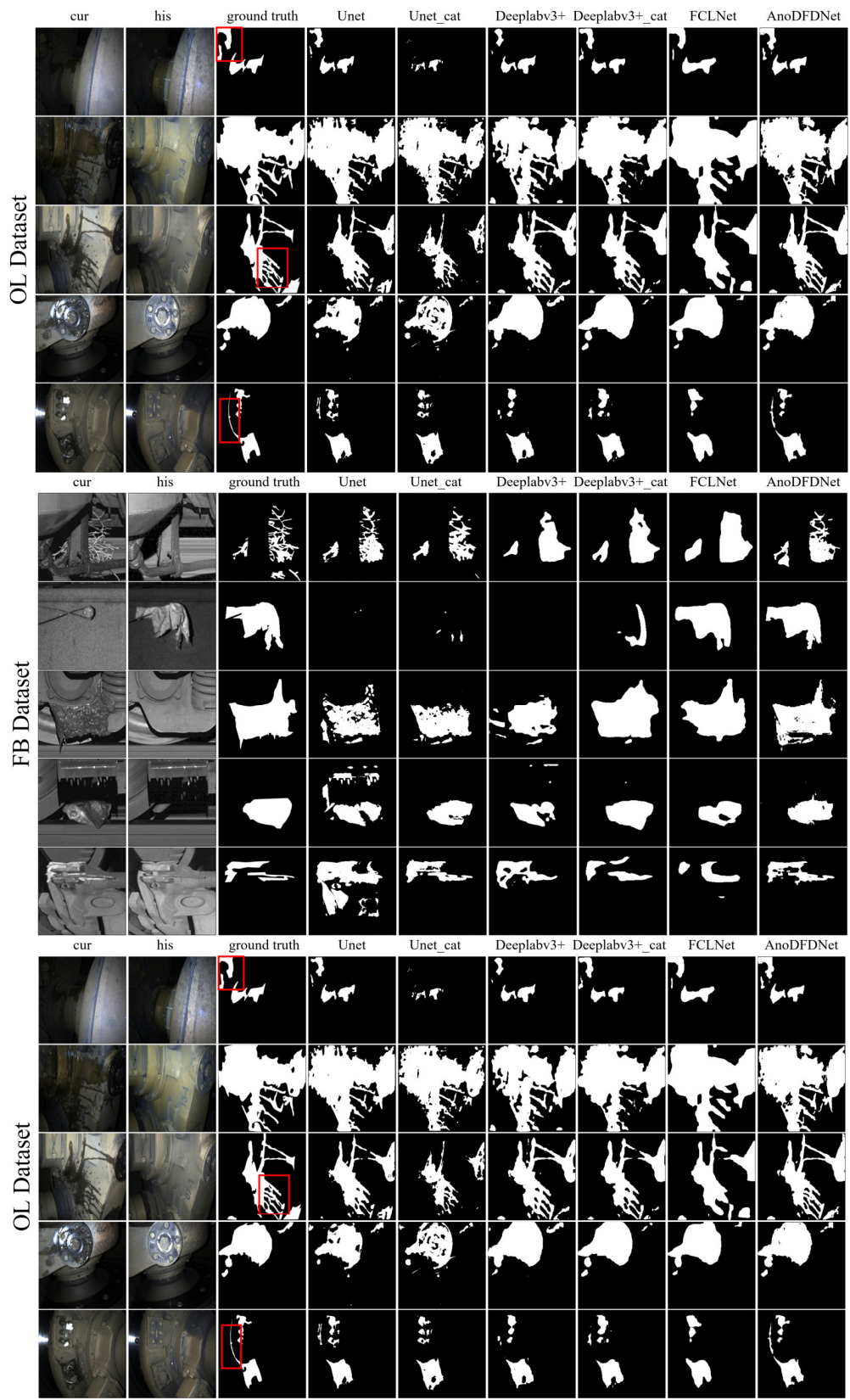

Fig. 6. Visualization of the anomaly maps of different methods on three datasets. From left to right in each row, the image denotes 'cur' image, 'his' images, ground truth, the anomaly map obtained by Unet, Unet-cat, Deeplabv3+, Deeplabv3+_cat, FCLNet, and our AnoDFDNet, respectively. Here the white pixel denotes detected anomies and the black denotes the background without anomalies.

TABLE III
RESULTS ON TRANSFORMER LAYER NUMBER.

| Datasets | No | Layer Number | Pre | Re | OA | F1 | IoU |
|---|---|---|---|---|---|---|---|
| Diff Dataset | 1 | 0 | 0.6473 | 0.7124 | 0.9920 | 0.6783 | 0.5132 |
|  | 2 | 1 | 0.7027 | **0.7585** | 0.9934 | 0.7296 | 0.5743 |
|  | 3 | 2 | 0.7880 | 0.7350 | 0.9945 | 0.7606 | 0.6136 |
|  | 4 | 4 | **0.7867** | 0.7138 | 0.9943 | 0.7485 | 0.5981 |
|  | 5 | 8 | 0.7735 | 0.7515 | **0.9944** | **0.7624** | **0.6160** |
| FB Dataset | 6 | 0 | 0.7952 | **0.7897** | 0.9319 | 0.7925 | 0.6563 |
|  | 7 | 1 | 0.8372 | 0.7731 | **0.9379** | 0.8038 | 0.6720 |
|  | 8 | 2 | 0.8328 | 0.7891 | 0.9392 | **0.8104** | **0.6812** |
|  | 9 | 4 | **0.8479** | 0.7562 | 0.9375 | 0.7994 | 0.6659 |
|  | 10 | 8 | 0.8191 | 0.7537 | 0.9320 | 0.7850 | 0.6461 |
| OL Dataset | 11 | 0 | 0.8742 | 0.7689 | 0.9605 | 0.8182 | 0.6923 |
|  | 12 | 1 | 0.8532 | 0.8256 | **0.9634** | **0.8392** | **0.7229** |
|  | 13 | 2 | 0.8395 | 0.8227 | 0.9613 | 0.8310 | 0.7109 |
|  | 14 | 4 | 0.7979 | 0.8637 | 0.9590 | 0.8295 | 0.7087 |
|  | 15 | 8 | 0.8235 | 0.8514 | 0.9617 | 0.8372 | 0.7180 |

### C. Ablation Studies

In the proposed AnoDFDNet, a significant module is the Transformer-based architecture. To investigate the influence of Transformer layers on model performance, we implemented a series of ablation experiments around the number of Transformer layers, here '0' denotes without Transformer module in AnoDFDNet.

As shown in Table III, on OL dataset, the results with different Transformer layers are similar; on FB dataset, the model with 2 layers of Transformer obtained the highest performance; on Diff dataset, the performance is superior to others when layer number is set to 8. More importantly, the results with or without Transformer are very different. For example, in terms of the IoU, experiment No5 improved by 10.28% when compared with No1; experiment No8 has an improvement of 2.49% over No6; experiment No12 is 3.06% higher than No11. Form above results, it denotes that the designed Transformer-based module indeed improve the detecting performance. To some extent, the performance improvement benefits from the Transformer-based module especially on datasets with large viewpoint difference. Because the Transformer is good at establishing global semantic relations and modeling long-range context which benefits to samples with large viewpoint difference. It is worth noting that, although experiment No7, No8, and No9 has superior performances than No6, experiment No10 has lower performance than No6 on FB dataset. One reasonable explanation is that the samples of FB dataset is not very enough to train a complicated Transformer-based model with 8 layers of Transformers.

The anomaly maps generated by AnoDFDNet without Transformer and with two layers of Transformers are listed in Fig. 7. From Fig. 7, it shows Transformer-based AnoDFDNet outperforms the one without Transformer. On Diff dataset, for the first and second pairs of samples, no-Transformer AnoDFDNet identified part of background into anomalies; for the third pair of samples, no-Transformer AnoDFDNet missed some anomalies. On FB dataset, different from no-Transformer AnoDFDNet which detected much false positive, the Transformer-based AnoDFDNet abtained good anomaly maps on three pairs of samples. On OL dataset, as to the first pair of samples, no-Transformer AnoDFDNet missed a large abnormal region. From above experiment results, it verifies the superiority of the resigned AnoDFDNet.

### D. Visualization of Feature Maps and Dimensionality Reduction

To interpret the proposed AnoDFDNet as a deeper level, we visualized the feature maps abstracted from the 'd2' layer (as shown in Fig. 2). In addition, the T-SNE (T-distributed Stochastic Neighbor Embedding) [41] algorithm is used to embed the feature distribution.

As shown in Fig. 8, we selected 3 typical feature maps for each pair of samples. The feature map 1 mainly focuses on the background and the abnormal region's boundary. Feature map 2 and 3 response for the anomaly itself. All of them allow the model focus training on more useful feature representations. The last column visualized the three-dimensional feature embedding, it shows the proposed model obtained a good distinction between abnormal and normal feature information.

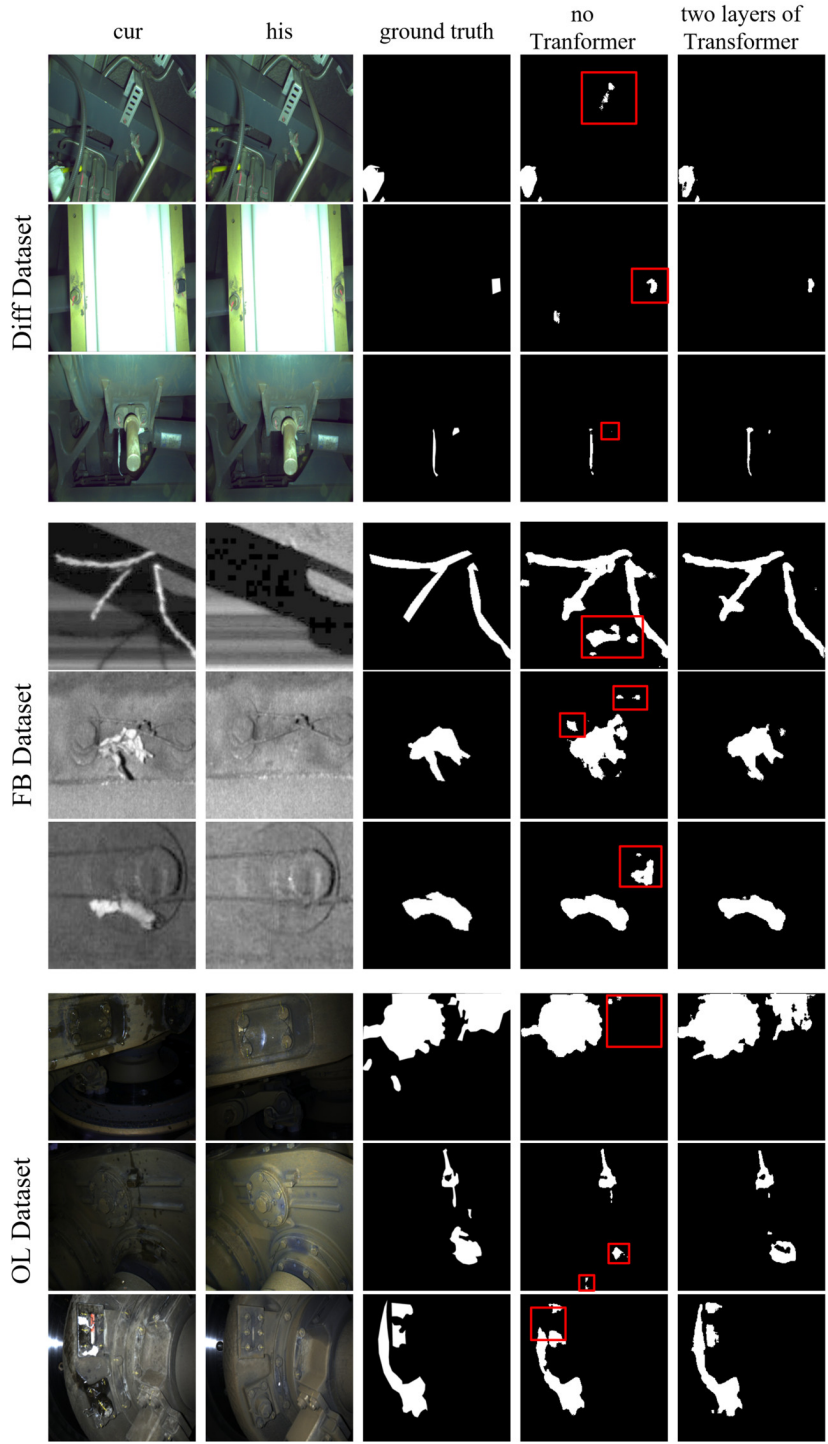

Fig. 7. Anomaly map visualization of AnoDFDNet without Transformer and with two layers of Transformers on three datasets. From left to right in each row, the image denotes 'cur' image, 'his' images, ground truth, the anomaly map obtained by AnoDFDNet without Transformer, the anomaly map obtained by AnoDFDNet with two layers of Transformers, respectively.

## VI. CONCLUSION

In this paper, we proposed a novel anomaly detection framework named AnoDFDNet based on the fusion of convolutional neural networks and the Vision Transformer. In view of the drawbacks of detecting anomalies with a single image, we introduce the idea that the 'anomaly' describes an abnormal state instead of a specific object. Therefore, it is more reasonable that the anomaly detection should be operated upon a pair of images. By detecting the interesting image difference, the abnormal information can be identified. Following above idea, we established a deep feature difference anomaly detection model, in which convolutional neural networks and the Vision Transformer are fused together. Extensive experiments indicate the proposed method outperforms other comparison methods. In addition, the integrated Transformer layers indeed improve detection performance and benefit to detect such samples with large viewpoint difference. In terms of the F1 score, the AnoDFDNet obtained 76.24%, 81.04%, and 83.92% on Diff dataset, FB dataset, and OL dataset, respectively. This paper provided a novel method of both researchers and practitioners to further develop anomaly detection task.

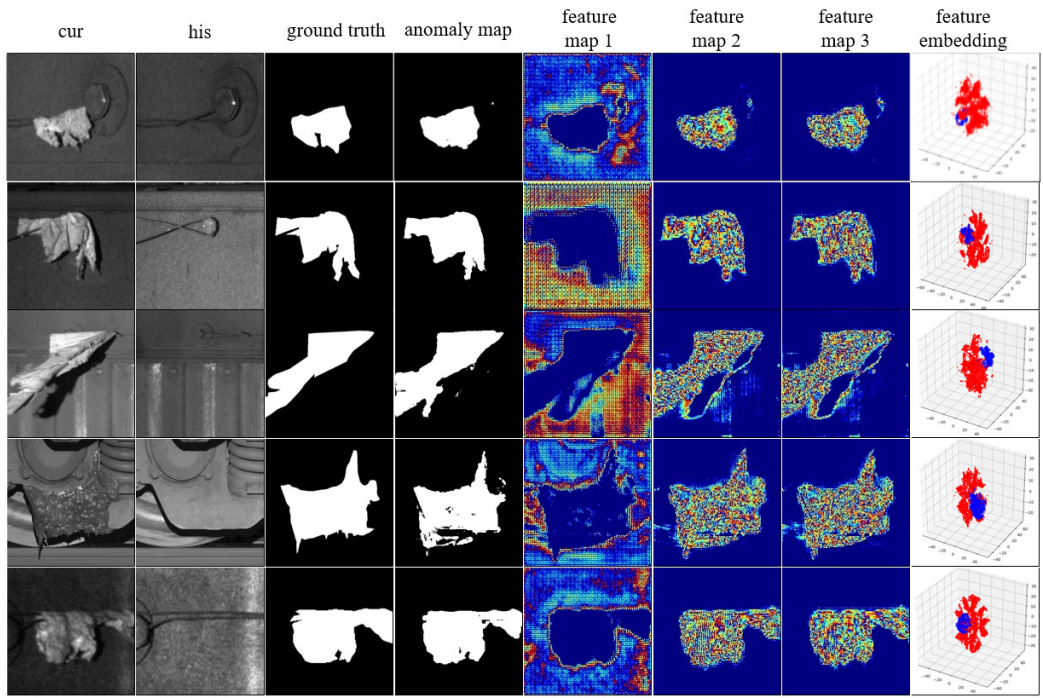

Fig. 8. Visualization of feature maps generated by layer 'd2' and its embedded feature representation on FB dataset. From left to right, the image denotes 'cur' image, 'his' images, ground truth, the anomaly map obtained by AnoDFDNet, three-dimensional feature embedding (where red denotes normal features and blue denotes abnormal features), respectively. In each feature map, read denotes high response and blue denotes low response. (Best in color)

REFERENCES

[1] C. Raghavendra, and S. Chawla, "Deep learning for anomaly detection: A survey," 2019, arXiv: 1901.03407. [Online]. Available: https://arxiv.org/abs/1901.03407
[2] G. S. Pang, C. H. Shen, L. B. Cao, and H. Ven, "Deep learning for anomaly detection: a review," ACM Comput. Surv., vol. 54, no. 2, pp. 1-38, Mar. 2021.
[3] L. Ruff, J. R. Kauffmann, R. A. Vandermeulen, G. Montavon, W. Samek, M. Kloft, T. G. Dietterich, and K. Muller, "A Unifying Review of Deep and Shallow Anomaly Detection," Proc. IEEE., vol. 109, no. 5, pp. 756-795, May. 2021.
[4] A. Aisha, M. Mohd Aizaini, and Z. Anazida, "Fraud detection system: a survey," J. Netw. Comput. Appl., vol. 68, pp. 90-113, Jun. 2016.
[5] X. M. Xie, C. Y. Wang, S. Chen, G. M. Shi, and Z. F. Zhao, "Real-time illegal parking detection system based on deep learning," in Proc. Int. Conf. Deep. Learn. Technol., Qingdao, China, 2017, pp. 23-27.
[6] K. B. Ravi, T. D. Mathew, and P. Ranjith, "An overview of deep learning-based methods for unsupervised and semi-supervised anomaly detection in videos," Journal of imaging., vol. 4, no. 2, pp. 1-36, Jan. 2018.
[7] T. Schlegl, P. Seebock, S. M. Waldstein, U. Schmidt-Erfurth, and G. Langs, "Unsupervised anomaly detection with generative adversarial networks to guide marker discovery," in Proc. Int. Conf. Inf. Proc. Med. Imaging. Cham, 2017, pp. 146-157.
[8] A. Samet, A. Amir, and B. Toby P, "GANomaly: semi-supervised anomaly detection via adversarial training," in Proc. 14th Asian. Conf. Comput. Vision., Australia, 2019, pp. 622-637.
[9] S. Mohammadreza, S. Niousha, B. Soroosh, R. Mohammad H, and R. Hamid R, "Multiresolution knowledge distillation for anomaly detection," in Proc. IEEE Conf. Comput. Vis. Pattern. Recognit., Long Beach, USA, 2021, pp. 14897-14907.
[10] B. Paul, F. Michael, S. David, and S. Carsten, "Uninformed students: student-teacher anomaly detection with discriminative latent embeddings," in Proc. IEEE Conf. Comput. Vis. Pattern. Recognit., Long Beach, USA, 2021, pp. 4182-4191.
[11] P. L. Mazzeo, M. Nitti, E. Stella, and A. Distante, "Visual recognition of fastening bolts for railroad maintenance," Pattern Recognit. Lett., vol. 25, no. 6, pp. 669-677, Apr. 2004.
[12] P. L. Mazzeo, E. Stella, and A. Distante, "Visual recognition of fastening bolt in railway maintenance context by using wavelet transform," International journal on graphics, vision and image processing., vol. Sl1, pp. 25-32, 2005.
[13] F. Marino, A. Distante, P. L. Mazzeo, and E. Stella, "A real-time visual inspection system for railway maintenance: Automatic hexagonal-headed bolts detection," IEEE Trans. Syst., vol. 37, no. 3, pp. 418–428, May. 2007.
[14] P. De Ruvo, A. Distante, E. Stella, and F. Marino, "A GPU-based vision system for real time detection of fastening elements in railway inspection," in Proc. 16th IEEE Int. Conf. Image. Proc., Cairo, Egypt, 2009, pp. 2333-2336.
[15] K. HyunCheo, and K. Whoi-Yul, "Automated inspection system for rolling stock brake shoes," IEEE Trans. Instrum. Meas., vol. 60, no. 8, pp. 2835-2847, Apr. 2011.
[16] L. Liu, F. Q. Zhou, and Y. Z. He, "Automated visual inspection system for bogie block key under complex freight train environment," IEEE Trans. Instrum. Meas., vol. 65, no. 1, pp. 2-14, Jan. 2016.
[17] J. H. Sun, and Z. W. Xiao, Y. X. Xie, "Automatic multi-fault recognition in TFDS based on convolutional neural network," Neurocomputing., vol. 222, pp. 127-136, Jan. 2017.

[18] J. H. Sun, and Z. W. Xiao, "Potential fault region detection in TFDS images based on convolutional neural network," in Proc. Infrared. Technol. Appl. Rot. Adv. Control., 2016, pp. 384-391.
[19] J. W. Chen, Z. G. Liu, H. R. Wang, N. Alfredo, and Z. W. Han, "Automatic defect detection of fasteners on the catenary support device using deep convolutional neural network," IEEE Trans. Instrum. Meas., vol. 67, no. 2, pp. 257-269, Feb. 2018.
[20] Y. He, K. Song, Q. Meng, and Y. Yan, "An end-to-end steel surface defect detection approach via fusing multiple hierarchical features," IEEE Trans. Instrum. Meas., vol. 69, no. 4, pp. 1493-1504, Apr. 2020.
[21] A. Q. Ma, Z. M. Lv, X. J. Chen, Y. J. Qiu, S. B. Zheng, and X. D. Chai, "Pandrol track fastener defect detection based on local convolutional neural networks," Proc. Ins. Mech. Eng. Part I-J Syst Control Eng., vol. 235, no. 10, pp. 1906-1915, Sep. 2020.
[22] X. K. Wei, Z. M. Yang, Y. X. Liu, D. H. Wei, L. M. Jia, and Y. J. Li, "Railway track fastener defect detection based on image processing and deep learning techniques: a comparative study," Eng. Appl. Artif. Intell., vol. 80, pp. 66-81, Jan. 2019.
[23] H. Feng, Z. G. Jiang, F. Y. Xie, P. Yang, J. Shi, and L. Chen, "Automatic fastener classification and defect detection in vision-based railway inspection systems," IEEE Trans. Instrum. Meas., vol. 3, no. 4, pp. 877-888, Oct. 2014.
[24] Z. Tu, S. Wu, G. Kang and J. Lin, "Real-Time defect detection of track components: considering class imbalance and subtle difference between classes," IEEE Trans. Instrum. Meas., vol. 70, pp. 1-12, Oct. 2021.
[25] Y. Zhan, X. X. Dai, E. H. Yang, and K. C. Wang, "Convolutional neural network for detecting railway fastener defects using a developed 3D laser system," Int. J. Rail Transp., vol. 9, no. 5, pp. 424-444, Sep. 2020.
[26] Z. X. Wang, J. P. Peng, W. W. Song, X. R. Gao, Y. Zhang, L. F. Xiao, and L. Ma, "A convolutional neural network-based classification and decision-making model for visible defect identification of high-speed train images," J. Sens., vol. 2021, no. 5554920, Mar. 2021.
[27] Y. He, J. Wu, Y. Zheng, Y. Zhang and X. Hong, "Track defect detection for high-speed maglev trains via deep learning" IEEE Trans. Instrum. Meas., vol. 71, pp. 1-8, Feb. 2022.
[28] L. Yao, J. Qiu, S. Gao, X. Du, Z. Gu and H. Song, "Defect detection in high-speed railway overhead contact system: importance, challenges, and methods," in Prof. Int. Conf. Secur. Pattern. Anal. Cybern., Chengdu, China, 2021, pp. 76-80.
[29] Y. Q. Bao, K. C. Song, Y. Y. Wang, Y. H. Yan, H. Yu, and X. J. Li, "Triplet-graph reasoning network for few-shot metal generic surface defect segmentation," IEEE Trans. Instrum. Meas., vol. 70, pp. 1-11, May. 2021.
[30] J. Cao, G. Yang and X. Yang, "A pixel-level segmentation convolutional neural network based on deep feature fusion for surface defect detection," IEEE Trans. Instrum. Meas., vol. 70, pp. 1-12, Dec. 2021.
[31] L. M. Li, R. Sun, S. G. Zhao, X. D. Chai, S. B. Zheng, and R. C. Shen, "Semantic-segmentation-based rail fastener state recognition algorithm," Math. Probl. Eng., vol. 8956164, May. 2021.
[32] Y. Ou, J. Q. Luo, B. L. Li, and B. He, "A classification model of railway fasteners based on computer vision," Neural Comput. Appl., vol. 31, no. 12, pp. 9307-9319, Jul. 2019.
[33] X. Giben, V. M. Patel and R. Chellappa, "Material classification and semantic segmentation of railway track images with deep convolutional neural networks," in Proc. IEEE Int. Conf. Image. Proc., Quebec City, Canada, pp. 621-625.
[34] Z. X. Wang, X. J. Tu, X. R. Gao, C. Y. Peng, L. Lin, and W. W. Song, "Bolt detection of key component for high-speed trains based on deep learning", in Proc. Far. East. Net. New. Technol. Appl. Forum., Qingdao, China, 2019, pp. 192-196.
[35] A. Dosovitskiy, L. Beyer, A. Kolesnikov, D. Weissenborn, X. Zhai, T. Unterthiner, M. Dehghani, M. Minderer, G. Heigold, S. Gelly, J. Uszkoreit, N. Houlsby, "An image is worth 16x16 words: Transformers for image recognition at scale," in Proc. 9th Int. Conf. Learn. Represent., Vienna. Austria, 2021.
[36] Z. X. Wang, Y. Zhang, L. Luo, and N. Wang, "TransCD: scene change detection via transformer-based architecture," Opt. Express., vol. 29, no. 25, pp. 41409-41427, Nov. 2021.
[37] S. Zheng, J. Lu, H. Zhao, X. Zhu, Z. Luo, Y. Wang, Y. Fu, J. Feng, T. Xiang, P. Torr, and L. Zhang, "Rethinking semantic segmentation from a sequence-to-sequence perspective with Transformers," in IEEE Conf. Comput. Vis. Pattern Recognit., 2021, pp. 6881-6890.
[38] O. Ronneberger, P. Fischer, and T. Brox, "U-Net: convolutional networks for biomedical image segmentation," in Proc. Int. Conf. Med. Image. Comput. Comput. -Assisted. Intervention., Munich, Germany, 2015, pp. 234-241.
[39] L. C. Chen, G. Papandreou, F. Schroff, and H. Adam, "Rethinking atrous convolution for semantic image segmentation," 2019, arXiv:1706.05587. [Oneline]. Available: https://arxiv.org/abs/1706.05587
[40] Z. X. Wang, C. Peng, Y. Zhang, N. Wang, and L. Luo, "Fully convolutional Siamese networks-based change detection for optical aerial images with focal contrastive loss," Neurocomputing., vol 457, pp. 155–167, Oct. 2021.
[41] L. V. D. Maaten and G. Hinton, "Visualizing data using t-sne," J. Mach. Learn. Res., vol. 9, pp. 2579–2605, Sep. 2008.